\documentclass[letterpaper, 10 pt, conference]{ieeeconf}  

\IEEEoverridecommandlockouts
\usepackage[dvipdfmx]{graphicx}
\usepackage{amsmath, amssymb,amsfonts}
\usepackage{ascmac,setspace}
\usepackage{cite}
\usepackage{color}

\title{\LARGE \bf
Effect of the Dynamics of a Horizontally Wobbling Mass\\ on Biped Walking Performance
}

\author{Tomoya Kamimura$^{1}$ and Akihito Sano$^{1}$
\thanks{$^{1}$Tomoya Kamimura and Akihito Sano are with Department of Electrical and Mechanical Engineering, Nagoya Institute of Technology, Nagoya, Japan
        {\tt\small kamimura.tomoya@nitech.ac.jp}}%
}

\begin{document}

\maketitle
\thispagestyle{empty}
\pagestyle{empty}

\begin{abstract}
We have developed biped robots with a passive dynamic walking mechanism. This study proposes a compass model with a wobbling mass connected to the upper body and oscillating in the horizontal direction to clarify the influence of the horizontal dynamics of the upper body on bipedal walking. The limit cycles of the model were numerically searched, and their stability and energy efficiency was investigated. Several qualitatively different limit cycles were obtained depending mainly on the spring constant that supports the wobbling mass. Specific types of solutions decreased the stability while reducing the risk of accidental falling and improving the energy efficiency. The obtained results were attributed to the wobbling mass moving in the opposite direction to the upper body, thereby preventing large changes in acceleration and deceleration while walking. The relationship between the locomotion of the proposed model and the actual biped robot and human gaits was investigated.
\end{abstract}

\section{Introduction}
Bipedal walking is generated through the dynamic interactions among the body, neural system, and environment.
McGeer~\cite{McGeer1990} developed a simple robot with passive legs attached to the hip and demonstrated stable walking while descending a slope without any energy input other than gravity.
This suggested that passive locomotion plays a significant role in locomotion.
Many studies have clarified the gait generation mechanism by using a simple compass model consisting of a body mass and two rigid bars as legs~\cite{Garcia1998,Goswami1998,Garcia2000,Collins2005,Obayashi2015,Okamoto2020}.
In addition to completely passive models, models with controls derived from physiological studies to imitate the neural system have also been proposed~\cite{Rybak2006,Cappellini2006,Ivanenko2007}.
For example, Aoi et al. \cite{Aoi2006,Aoi2007} applied control based on a central pattern generator (CPG), which involves a rhythm resetting (i.e., phase resetting) function of human neural systems, to the compass model, and revealed the mechanism through which stable limit cycles are generated.

The dynamics of both the lower limbs and the upper body play an important role in walking.
Therefore, many researchers have investigated the effect of the upper body on locomotion.
For example, in modeling studies, a rigid upper body has been attached to simple models such as a compass model, and its effect on gait has been investigated~\cite{Wisse2004,Wisse2007,Collins2009,Ackerman2013,Ackerman2014,Honjo2013,Honjo2019}.
Notably, the upper body is not a single rigid body but consists of multiple soft tissues.
Therefore, many researchers have focused on the active or passive wobbling mass of the upper body, the flexible spine, and the pendulum-like oscillation of the arms~\cite{Rome2006,Arellano2014,Hanazawa2015, Hanazawa2019, Toda2020}. These studies showed that such elements compensated for the torque generated by legs, reduced the peak ground reaction force, increased the energy efficiency, and increased or decreased the stability.
In our previous study, we showed that a passive wobbling mass connected to the upper body and oscillating in the vertical direction generates human-like time profiles of the ground reaction force while running using a simple model and a running biped robot~\cite{Kamimura2021}.

While the vertical wobbling mass plays a significant role in running, as shown in our previous study, a horizontal wobbling mass is also expected to play an important role in walking.
For example, the horizontal wobbling mass is assumed to affect acceleration and deceleration in the horizontal direction and thus impact the energy efficiency of gait.
Moreover, the horizontal movement of the upper body is supposed to affect the stability.
During walking, the zero moment point (ZMP) always exists inside the support polygon~\cite{Popovic2005}.
If the ZMP moves outside the support polygon, the walker will fall down.
The horizontal movement of the body is assumed to play a significant role in the dynamics of the ZMP.
Therefore, in this study, we investigate the effect of the horizontal dynamics of the upper body on the walking performance in terms of the energy efficiency and stability.
To clarify this effect, based on a compass model, we propose a simple walking model with a horizontally oscillating mass point (i.e., a wobbling mass).
Furthermore, we discuss the relationship between our simple model and actual walking.

\section{Model}
\begin{figure}
    \centering
    \includegraphics[scale=0.9]{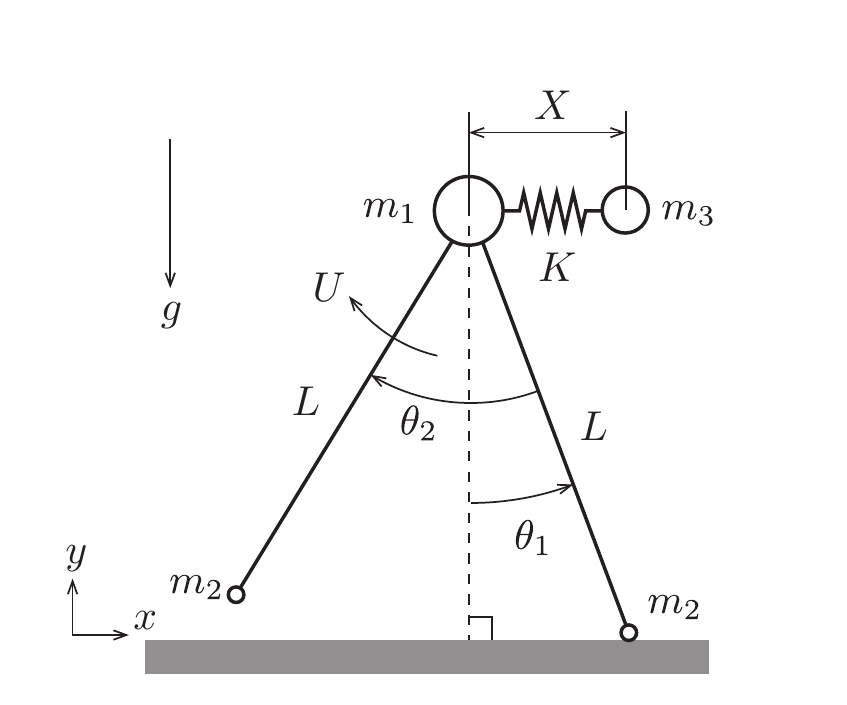}
    \caption{Compass model with elastically supported horizontal wobbling mass.}
    \label{fig:model}
\end{figure}
To investigate the dynamical effect of the upper body on walking using a minimal model, we proposed a model based on a compass model equipped with a wobbling mass, as shown in Fig.~\ref{fig:model}.
The model consists of a hip, swing leg, stance leg, and wobbling mass.
The horizontally wobbling mass connected to the body through a spring represents the interaction between the the lower and upper body in horizontal direction.
The legs have length $L$ and are connected at the hip.
Each foot can detect the touchdown timing on the ground.
The tip of the stance leg is constrained on the ground and it behaves as a frictionless pin joint.
The body mass $m_1$ and leg mass $m_2$ are assumed to be concentrated at the hip and tip, respectively.
The wobbling mass $m_3$ is connected to the hip through a prismatic spring $K$.
It can move only in the horizontal direction, and its height always equals the hip height.
The whole upper body mass is $M = m_1 + m_3$, where $m_3 = \alpha M$ and $m_1 = (1-\alpha)M$.
The model is constrained on the $x$-$y$ plane, with the walking direction along the $x$-axis.
The model has three degrees of freedom, $\theta_1$, $\theta_2$, and $X$, where $\theta_1$ is the angle of the stance leg with respect to the vertical direction, $\theta_2$ is the angle between the swing leg and the stance leg, and $X$ is the distance between the hip and the wobbling mass.
In this model, $\theta_1$ and $X$ are not controlled directly, whereas $\theta_2$ is controlled by the actuator torque $U$.

We derived the dimensionless governing equations using the characteristic length $L$ and characteristic time scale $\tau = \sqrt{L/g}$.
The state variables of the model are defined as $q = [\theta_1,\theta_2,x]^\top$, where $x = X/L$.
The dimensionless equations of motion for the swing phase are given by
\begin{align}
    \mathcal{M}(q)\ddot{q} + h(q,\dot{q}) + v(q) = Q,
\end{align}
where $\mathcal{M}(q)$ is the inertia matrix, $h(q,\dot{q})$ is the Coriolis and centrifugal forces, $v(q)$ is the conservative force, and $Q = [0,u = U/MgL,0]^\top$ is the input torque.
$\dot{*}$ and $\ddot{*}$ respectively indicate the first and second derivatives of variable $*$ with respect to $t/\tau$.

The swing leg touches the ground when $2\theta_1 = \theta_2$ is satisfied, at which time the swing and stance legs are immediately switched.
The impulsive force occurs at the tip and results in a discontinuous change in the velocities.
From the law of conservation of angular momentum, the relationship between the states immediately before and after touch down is denoted using function $H$ as
\begin{align}
    [(q^+)^\top, (\dot{q}^+)^\top]^\top = H(q^-,\dot{q}^-),
\end{align}
where $*^-$ and $*^+$ indicate the states immediately before and after touchdown, respectively.

The input torque $u$ is determined based on a rhythmic signal oscillator (CPG)~\cite{Aoi2006}
\begin{align}
    u = -K_{\rm p}(\theta_2 - \theta_2^{\rm d}(\gamma,\phi)) - K_{\rm d}(\dot{\theta}_2 - \dot{\theta}_2^{\rm d}(\gamma,\phi)),
\end{align}
where the desired angle $\theta_2^{\rm d}(\gamma,\phi))$ is defined as
\begin{align}
    \theta_2^{\rm d}(\gamma,\phi) = \gamma(1+ \cos \phi) - S,
\end{align}
where $\gamma$ and $\phi$ are respectively the amplitude and phase of the oscillator, which has constant angular velocity $\omega$.
Further, $S$ is the stride angle.

\subsection{Searching limit cycles}
We defined the Poincar\'{e} section at the touchdown moment.
A Poincar\'{e} map $P$ was defined as
\begin{align}
    z_{n+1} = P(z_n),
\end{align}
where $z_n$ is the state variable at the $n$th intersection with the Poincar\'{e} section.
For a periodic gait, $z^* = P(z^*)$ is satisfied, where $z^*$ is a fixed point on the Poincar\'{e} section.
We numerically searched for fixed points for periodic walking by using the \texttt{fsolve} function in MATLAB.

\subsection{Stability and risk of accidental falling}
We used the eigenvalues of the linearized Poincar\'{e} map around the fixed point on the Poincar\'{e} section.
The limit cycle is stable when all of the eigenvalues are inside the unit cycle in the complex plane (these magnitudes are less than 1).

The ZMP is also an important measure to determine whether walking is stable or not.
Because the moment is zero around the ZMP, the horizontal position $p$ of the ZMP is given by
\begin{align}
    p = x_g - \frac{\ddot{x}_g}{g+\ddot{y}_g}y_g,
\end{align}
where $(x_g, y_g)$ are respectively the horizontal and vertical positions of the center of mass of the whole body, including the wobbling mass.
Generally, the ZMP is equal to the Center of Pressure (CoP) on the foot, however, in this study, the ZMP is defined as the point at which the moment of the apparent gravity, including acceleration, becomes zero. In this case, the position of ZMP and CoP do not necessarily coincide.
Although our model does not have a support polygon because it has point tips, we used the distance $d$ between the ground point and the ZMP as a criterion of the risk of accidental falling while walking.

\subsection{Energy efficiency}
To evaluate the energy efficiency, we defined the cost of transport (CoT) as
\begin{align}
    {\rm CoT} = \frac{W}{mg\bar{v}}
\end{align}
where $W$ is the work generated by the hip actuator during one stride, and $\bar{v}$ is the averaged horizontal velocity of the center of mass for one gait cycle.
A smaller CoT indicates better energy efficiency because the actuator expends lesser energy in one stride.

\section{Results}

\subsection{Obtained solution groups}
\begin{figure}
    \centering
    \includegraphics[width=80mm]{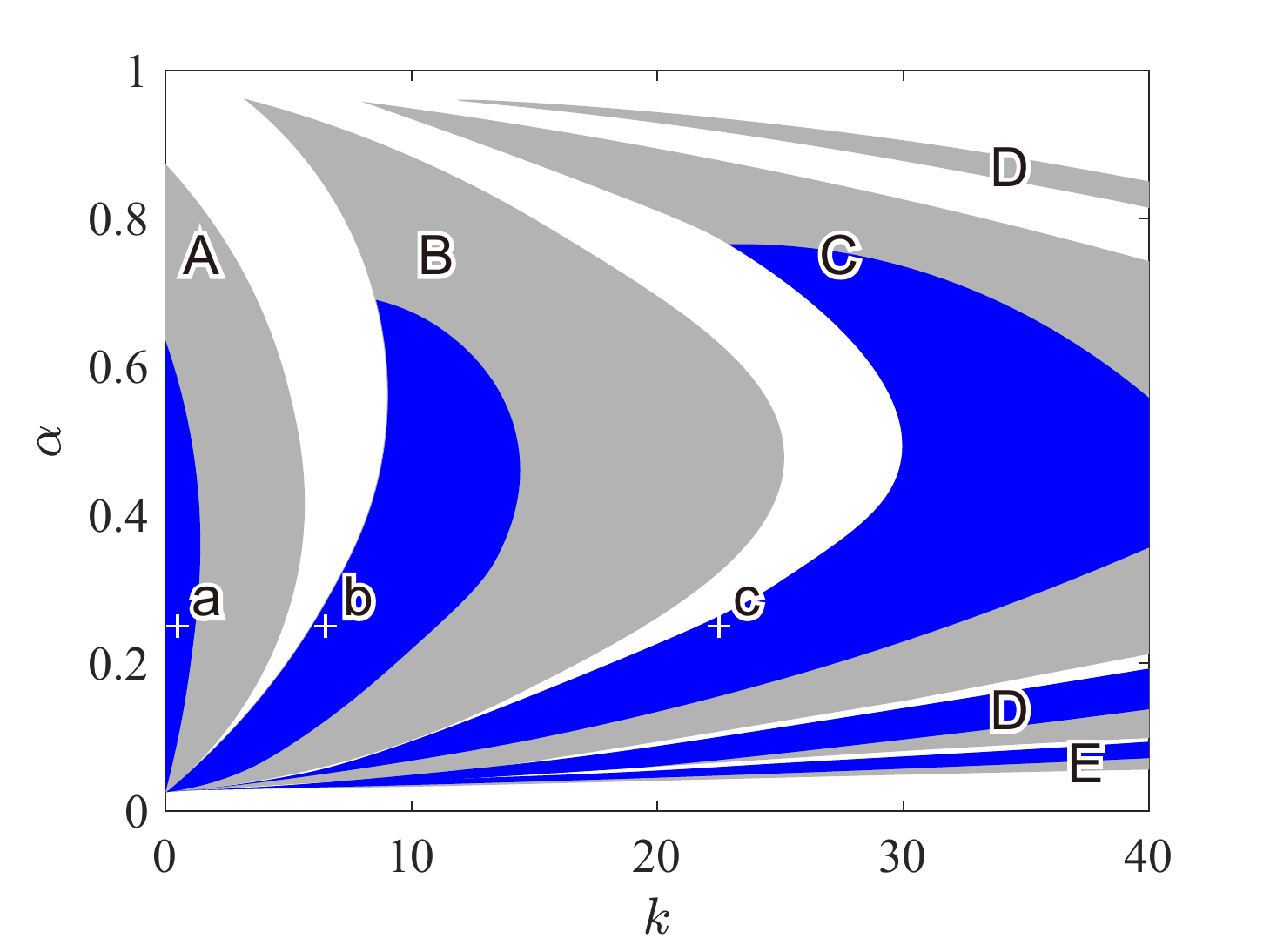}
    \caption{Distribution of limit cycles on $k$-$\alpha$ plane with $\omega = 3$. Stable and unstable limit cycles were found in blue and grey areas, respectively. Solution was not found in the white area. Solutions exhibit qualitatively different behaviors and are accordingly classified into groups A, B, C, D, and E. Points a, b, and c indicate typical stable solutions, whose behaviors are shown in Fig.~\ref{fig:variables}.}
    \label{fig:soltuons_distribution}
\end{figure}

The periodic solutions obtained for various $k$ and $\alpha$ values using the CPG frequency $\omega = 3$ are shown in Fig.~\ref{fig:soltuons_distribution}.
The obtained solutions are divided into several qualitatively different discrete groups; these are denoted as solution groups A, B, C, D, and E according to the value ranges of $k$, as shown in Fig.~\ref{fig:soltuons_distribution}.
Regardless of the $\omega$ value, the obtained solutions were classified into similar groups.

\begin{figure}
    \centering
    \includegraphics[width=80mm]{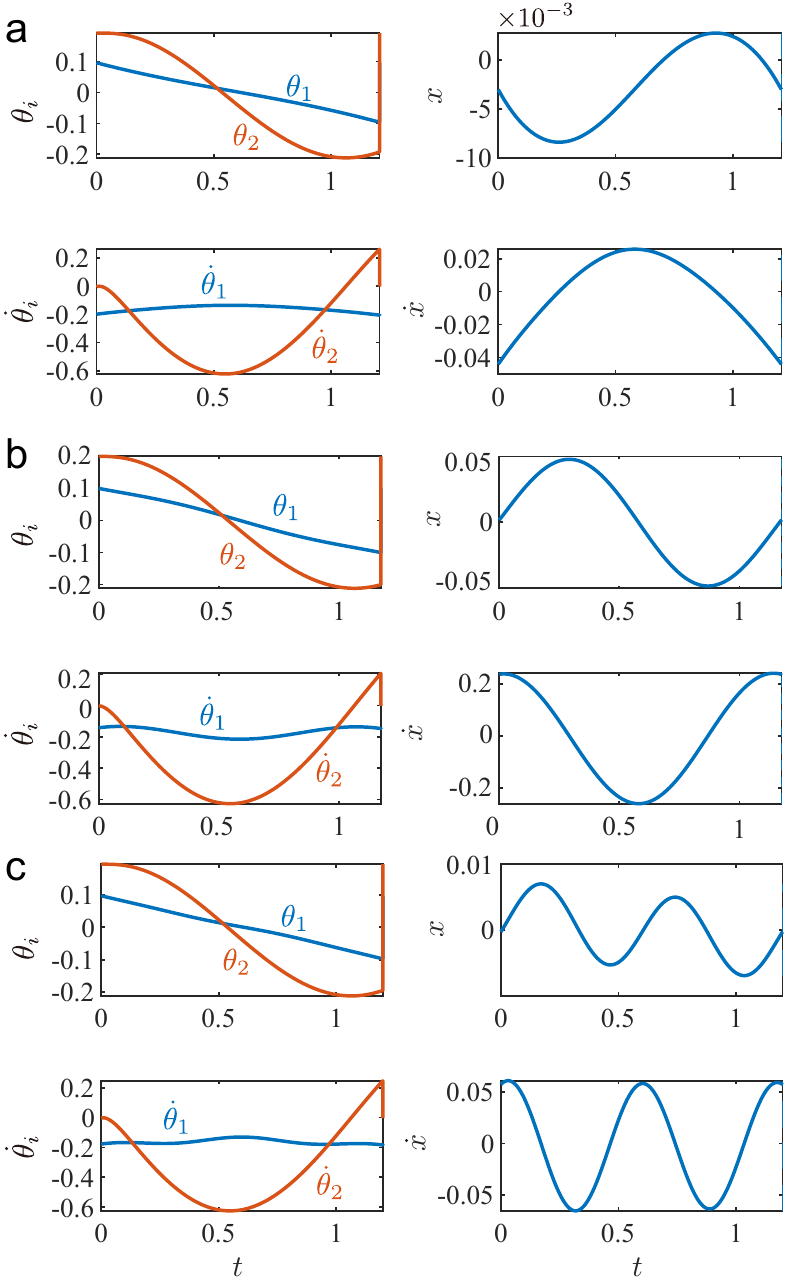}
    \caption{Time profiles of state variables of limit cycles of proposed model for three typical solutions a, b, and c in Fig.\ref{fig:soltuons_distribution}. (a) $k = 0.5$ (solution a), (b) $k = 6.0$ (solution b), and (c) $k = 22.5$ (solution c) for $\alpha = 0.25$ and $\omega = 3$.}
    \label{fig:variables}
\end{figure}

Figure~\ref{fig:variables} shows the behaviors of typical solutions of groups A, B, and C.
The solutions exhibited qualitatively different behaviors, particularly in $x$ and $\dot{\theta}_1$.
In the solutions of group A (Fig.~\ref{fig:variables}a), the position of the wobbling mass $x$ moves backward ($x < 0$) in the first half of the stance phase and forward ($x > 0$) in the second half.
By contrast, in the solutions of group B (Fig.~\ref{fig:variables}b), $x$ moves forward in the first half of the stance phase and backward in the second half.
In the solutions of group C (Fig.~\ref{fig:variables}c), $x$ oscillates two times during one gait cycle; this is regarded as a doubling of the oscillation of $x$ in the solutions of group B.
Similarly, in the solutions of groups D and E, $x$ oscillates three and four times, respectively.
Furthermore, $\dot{\theta}_1$ has one, two, and three peaks in the solutions of groups A, B, and C, respectively.

\begin{figure}
    \centering
    \includegraphics[width=80mm]{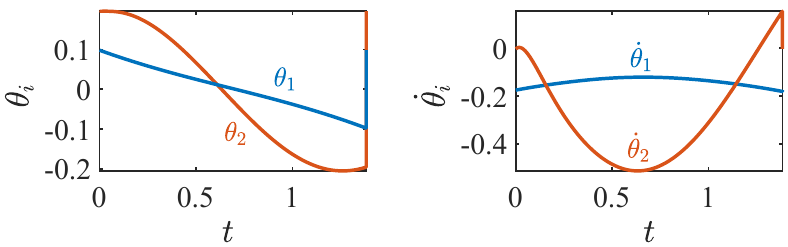}
    \caption{Time profiles of state variables of limit cycle of traditional compass model using same controller as that of the proposed model with $\omega = 3$.}
    \label{fig:variables_compass}
\end{figure}

The time responses of the state variables in limit cycles in the traditional compass model, which does not include a wobbling mass, applied with the same controller ($\omega = 3$) are shown in Fig.~\ref{fig:variables_compass}.
The behavior of the solution is similar to that of the solutions of group A of the proposed model because $\dot{\theta}_1$ has only one peak in one gait cycle.
The state variables $\theta_i, \dot{\theta}_i$ ($i = 1,2$) show approximately the same trajectory as that of the solutions of group A.
No other qualitatively different solutions were obtained for the compass model without the wobbling mass.

\subsection{Walking performances}
We evaluated the walking performances of the obtained solutions for various $k$, $\alpha$, and $\omega$ values.

\begin{figure}
    \centering
    \includegraphics[width = 80mm]{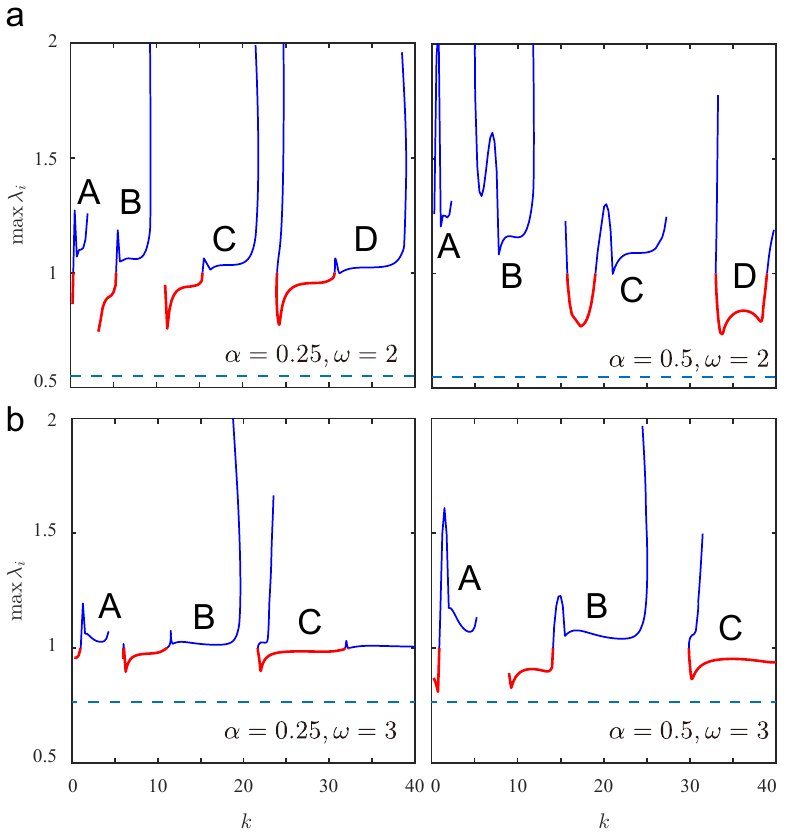}
    \caption{Maximum eigenvalues of obtained solutions for $\alpha = 0.25$ (left) and $\alpha = 0.5$ (right) for (a) $\omega = 2$ and (b) $\omega = 3$. A, B, C, and D indicate solution groups in Fig.~\ref{fig:soltuons_distribution}. Red and blue curves indicate stable and unstable solutions, respectively. Horizontal dashed lines indicate the maximum eigenvalue of the compass model without a wobbling mass.}
    \label{fig:eigenValues}
\end{figure}

Figure~\ref{fig:eigenValues} shows the maximum eigenvalues of the obtained limit cycles for various parameters.
In a single solution group, $\max \lambda_i$ takes the minimum value with small $k$ (Fig.~\ref{fig:soltuons_distribution}), which indicates smaller $k$ tends to result in higher stability.
Further, the solutions become unstable when $\alpha$ is too large.
When $\omega$ is large, the range of stable $k$ widens.
In comparison with the stability of the traditional compass model, the maximum eigenvalues of all solutions of the proposed model are larger (Fig.~\ref{fig:eigenValues}), indicating that the stability of the limit cycles is reduced by the effect of the wobbling mass.

\begin{figure}
    \centering
    \includegraphics[scale=1.0]{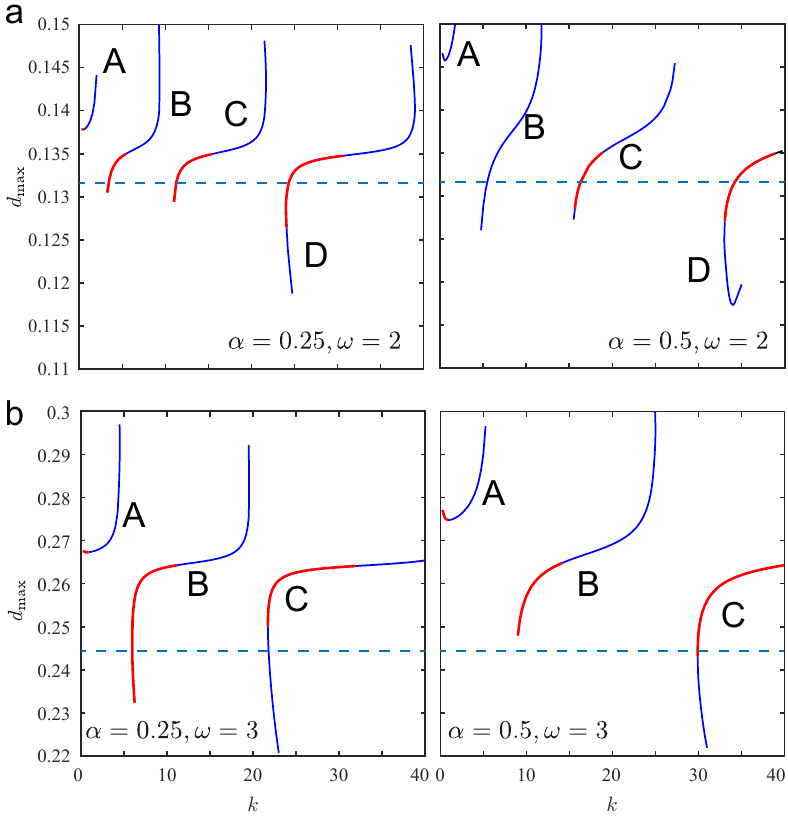}
    \caption{Distance $d$ between toe and ZMP of obtained solutions for $\alpha = 0.25$ (left) and $\alpha = 0.5$ (right) for (a) $\omega = 2$ and (b) $\omega = 3$. A, B, C, and D indicate solution groups in Fig.~\ref{fig:soltuons_distribution}. Red and blue curves indicate stable and unstable solutions, respectively. Horizontal dashed lines indicate $d_{\rm max}$ of the compass model without a wobbling mass.}
    \label{fig:ZMP}
\end{figure}

Figure~\ref{fig:ZMP} shows the distance $d_{\rm max}$, which indicates the maximum distance $d$ between the toe and ZMP for various parameters.
$d_{\rm max}$ is small for ($k,\alpha$), where the stability of the solutions is high (maximum eigenvalue is small).
The minimum value of $d_{\rm max}$ in the solutions of group A is larger than that in the solutions of groups B and C.
Furthermore, $d_{\rm max}$ is larger when $\alpha$ and $\omega$ are large.
In the solutions of group A, $d_{\rm max}$ is always larger than that of the traditional compass model.
However, in some stable solutions of groups B and C at $\omega = 2$, $d_{\rm max}$ is smaller than that of the traditional compass model.
In addition, $d_{\rm max}$ decreases rapidly with decreasing $k$.

\begin{figure}
    \centering
    \includegraphics[width = 80mm]{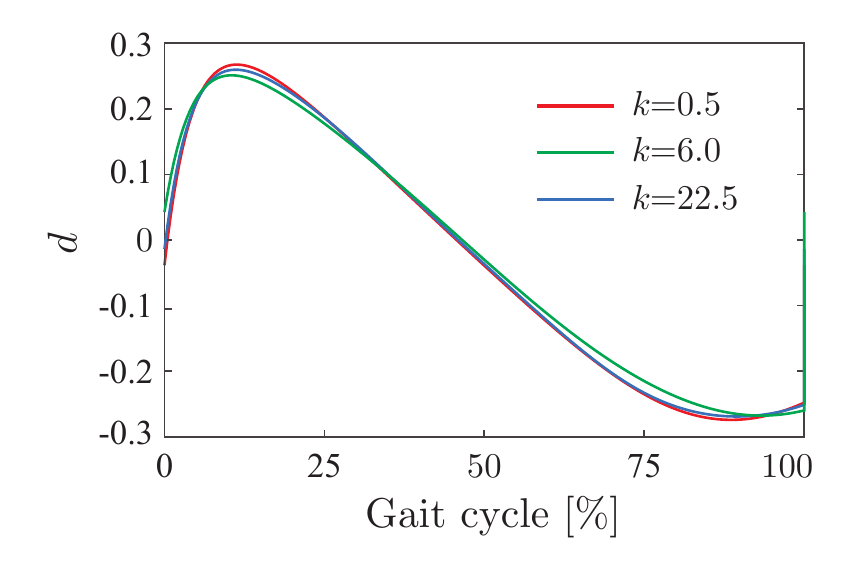}
    \caption{Time profiles of $d$ for $k = 0.5$ (red, solution a in Fig.~\ref{fig:soltuons_distribution}), $k = 6.0$ (green, solution b in Fig.~\ref{fig:soltuons_distribution}), and $k = 22.5$ (blue, solution c in Fig.~\ref{fig:soltuons_distribution}) for $\alpha = 0.25$, $\omega = 3$.}
    \label{fig:time-d}
\end{figure}

Moreover, the time profiles of $d$ for three typical stable solutions in groups A, B, and C with $\alpha = 0.25$ and $\omega = 3$ are shown in Fig.~\ref{fig:time-d}.
$d$ becomes positive in the first half of the stance phase and then slowly becomes negative.
Although there is no qualitative difference in the time profiles of $d$ among the solution groups, the solution with $k = 0.5$, which is a solution of group A, exhibits the largest fluctuation.

\begin{figure}
    \centering
    \includegraphics[scale=1.0]{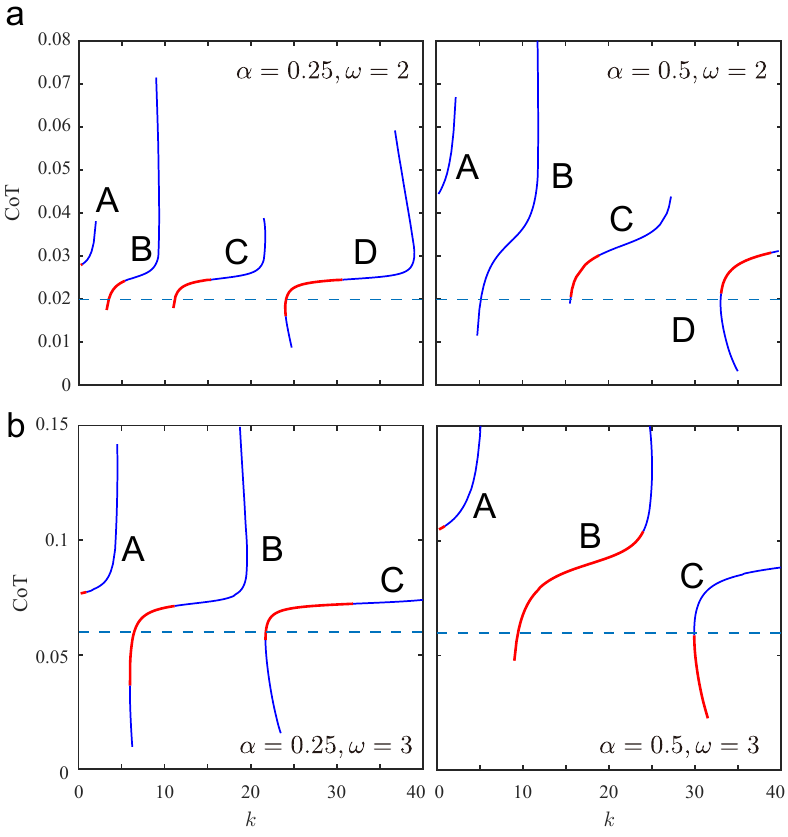}
    \caption{Energy efficiency (CoT) of obtained solutions for $\alpha = 0.25$ (left) and $\alpha = 0.5$ (right) for (a) $\omega = 2$ and (b) $\omega = 3$. A, B, C, and D indicate solution groups in Fig.~\ref{fig:soltuons_distribution}. Red and blue curves indicate stable and unstable solutions, respectively. Horizontal dashed lines indicate the CoT of the compass model without a wobbling mass.}
    \label{fig:CoT}
\end{figure}

The CoT for various parameters is shown in Fig.~\ref{fig:CoT}.
The smaller the CoT, the higher is the energy efficiency.
The CoT is small for ($k,\alpha$), where the limit cycles show high stability.
The minimum CoT value for solutions of group A is larger than that for solutions of groups B and C.
Furthermore, for larger $\alpha$ and $\omega$, the CoT is larger, indicating that the energy efficiency is lower.
Although the CoT of the proposed model is always larger than that of the compass model for solutions of group A, it is smaller than that of the compass model for some stable solutions of groups B and C.
Further, the CoT decreases rapidly as $k$ decreases.

\section{Discussion}

\subsection{Parameter dependency of the model}
The obtained results revealed that the body spring constant $k$ plays an important role in determining the solution group of the proposed compass model equipped with a wobbling mass.
As shown in Fig.~\ref{fig:soltuons_distribution}, depending on the spring constant, the solutions were obtained as some discrete groups (A, B, C, D, and E).
Solutions in different groups exhibit qualitatively different behaviors, particularly in the time profiles of the wobbling mass position $x$.

In usual walking, the body receives a deceleration force from the ground in the first half of the stance phase and an acceleration force in the second half of the stance phase.
In the solutions of group A, the wobbling mass moves backward in the first half of the stance phase and forward in the second half, as shown in Fig.~\ref{fig:variables}a.
In other words, it decelerates the body in the first half of the stance phase and accelerates the body in the second half.
By contrast, in the solutions of group B, the wobbling mass moves forward in the first half of the stance phase and backward in the second half, exerting force in a direction that cancels the acceleration/deceleration of the body.
Therefore, the large change in acceleration and deceleration of the body is suppressed by the wobbling mass in the solutions of group B.

The existence of solution groups is assumed to be associated with the natural frequencies of the body spring.
The solutions of group C have double the oscillation of the body spring than the solutions of group B.
When the spring constant is further increased, the solutions of groups D and E are seen to involve three- and four-fold oscillations, respectively.
Because the solutions are distributed in the $k$-$\alpha$ plane with a repetitive structure, solutions with $n$-fold period are assumed to exist when the spring constant is further increased.

Within a single group of solutions, a spring constant dependence was also observed.
The smaller the spring constant and the more stable the solution, the better is the energy efficiency and the closer is the ZMP to the toes.
When the solution is unstable, the actuator consumes a large amount of energy and undergoes large acceleration and deceleration, resulting in large ZMP fluctuations, poor energy efficiency, and poor stability.

The mass ratio $\alpha$ was also shown to affect the walking performance.
If $\alpha$ is too large or too small, few stable solutions are found, and the optimal value is approximately $0.4 < \alpha < 0.5$ (Fig.~\ref{fig:soltuons_distribution}).
By contrast, the larger the $\alpha$ value, the larger is the maximum distance $d_{\rm max}$ between the ZMP and the toes and CoT (Figs.~\ref{fig:ZMP} and~\ref{fig:CoT}).
The results indicate that $\alpha$ should not be too large when designing a robot with a wobbling mass.

Furthermore, as the phase angular velocity $\omega$ of the CPG increases, stable solutions are obtained over a wider range of $k$ and $\alpha$ values (Fig.~\ref{fig:eigenValues}).
However, CoT and $d$ also increase as $\omega$ increases (Figs.~\ref{fig:ZMP} and~\ref{fig:CoT}).
Walking at a fast pace requires large amounts of energy, resulting in large acceleration, deceleration, and $d$.

\subsection{Effect of wobbling mass}
We obtained several qualitatively different solutions for the proposed model with a wobbling mass.
The state variables in the limit cycles of solutions of group A showed a behavior similar to that of the traditional compass model (Figs.~\ref{fig:variables} and~\ref{fig:variables_compass}).
By contrast, solutions of groups B and C exhibited behaviors different from those of the traditional compass model, with $\dot{\theta}_1$ having multiple peaks and the wobbling mass undergoing oscillations.
This behavior arose owing to the effect of the body spring that supports the wobbling mass.

Furthermore, solutions of groups B and C had lower stability but better energy efficiency and a shorter distance between the ZMP and toe position than those of the traditional compass model.
The stability of a limit cycle is a measure of how quickly it converges when a perturbation is applied in the limit cycle.
The distance between the ZMP and the toes is a measure of whether or not a walker falls.
In solutions of groups B and C, the motion of the ZMP was suppressed by the motion of the wobbling mass.
Therefore, the risk of accidental falling while walking is expected to be reduced by the wobbling mass.
In addition, because the acceleration and deceleration of the body are suppressed to be small, the work exerted by the actuators is also small; this is thought to improve the energy efficiency.
These results indicate that the solutions of groups B and C use the wobbling mass to increase the energy efficiency and reduce the risk of accidental falling while walking.

The findings from the proposed model can be applied to biped walking robots. To prevent accidental falling, a passive horizontally wobbling mass can be equipped on the upper body of the biped robot. Because the attached wobbling mass requires no input energy, the stability of the robot can be improved with high energy efficiency as long as the gait period and spring stiffness are properly determined.

\subsection{Relationship between model and actual walking}
In this study, we investigated the effect of the dynamics of a horizontally wobbling mass on the walking performance.
The results showed that passively oscillating elements of the upper body could reduce the risk of accidental falling and improve the energy efficiency.
Therefore, the walking performance can be improved by attaching a horizontal wobbling mass to an actual walking biped robot.

In human walking, the soft tissues of the upper body may behave as a wobbling mass.
In the model solution, the displacement of the horizontal wobbling mass was approximately $|x| = 0.05$; in other words, this displacement was only approximately 5\% of the leg length.
Because even such a small oscillation can affect the gait, soft tissues in humans may play the same role as the wobbling mass in our model.
Furthermore, the results suggested the possibility of a passive walking assist device that oscillates back and forth.

Some birds tend to swing their heads forward and backward while walking.
Although birds reportedly perform such movements to stabilize their vision, it may have dynamic effects similar to those of the wobbling mass in the proposed model.
These movements may also have the effect of increasing the efficiency and stability of gait.

\section{Conclusion}
This study proposes a compass model with a wobbling mass that oscillates in the horizontal direction to reveal the influence of the horizontal motion of the upper body on walking.
We searched the limit cycles of the model and their stability, energy efficiency, and risk of falling were investigated.
Several qualitatively different limit cycles were obtained depending mainly on the spring constant that supports the wobbling mass.
Specific types of solutions reduced the risk of falling and improved the energy efficiency, although the stability was decreased compared to those of the traditional compass model.
Such results were due to the wobbling mass moving in the opposite direction to the upper body, thereby preventing large changes in acceleration and deceleration while walking.
The obtained results also suggest that humans can use the soft tissues of the upper body to improve the gait performance.

\subsection{Limitations and future works}
The proposed wobbling mass can move only in the horizontal direction. However, the human body has wobbling parts, including arms and internal organs, that can move like a pendulum or move vertically.
Future works will aim to clarify the relationship between these different oscillating parts and the actual human body.

We used a controller using a CPG for the proposed model, however, its effect has not been investigated except for the frequency.
We would like to investigate the effect of the neural system on the body with a wobbling mass.

Moreover, we would like to verify the validity of our results through robot experiments.
Furthermore, the gait of the proposed model will be compared with that of humans, and changes in gait when the walking support device is actually attached to the human body will be investigated.
Finally, the relationship with the gait of birds will be investigated, and the similarities and differences in the bipedal gait of humans and birds will be clarified.

\section*{Acknowledgement}
This work was supported in part by JSPS KAKENHI Grant Number JP21K14104 and JP22H01445.

\small
\bibliographystyle{IEEEtran}
\bibliography{ICRA2023}
\normalsize

\end{document}